\documentclass[11pt]{article}

\usepackage[final]{acl}

\usepackage{times}
\usepackage{latexsym}
\usepackage{amsmath}

\usepackage[T1]{fontenc}
\usepackage[utf8]{inputenc}

\usepackage{microtype}

\usepackage{inconsolata}

\usepackage{graphicx}

\title{Enhancing Robustness of Autoregressive Language Models against Orthographic Attacks via Pixel-based Approach}

\author{
 \textbf{Han Yang\textsuperscript{1,2}},
 \textbf{Jian Lan\textsuperscript{1,4}},
 \textbf{Yihong Liu\textsuperscript{3,4}},
 \textbf{Hinrich Schütze\textsuperscript{3,4}},
 \textbf{Thomas Seidl\textsuperscript{1,4}}
\\
\\
 \textsuperscript{1}LMU Munich, Munich, Germany\\
 \textsuperscript{2}GESIS - Leibniz Institute for the Social Sciences, Cologne, Germany\\
 \textsuperscript{3}Center for Information and Language Processing, LMU Munich, Germany\\
 \textsuperscript{4}Munich Center for Machine Learning (MCML), Germany\\
\texttt{han.yang@gesis.org} ~~~~~~
\texttt{\{lan, seidl\}@dbs.ifi.lmu.de}
}

\begin{document}
\maketitle
\begin{abstract}
Autoregressive language models are vulnerable to orthographic attacks, 
where input text is perturbed with characters from multilingual alphabets, 
leading to substantial performance degradation. 
This vulnerability primarily stems from the out-of-vocabulary issue inherent in subword tokenizers and their embeddings. 
To address this limitation, we propose a pixel-based generative language model that replaces the text-based embeddings with pixel-based representations by rendering words as individual images. 
This design provides stronger robustness to noisy inputs, while an extension of compatibility to multilingual text across diverse writing systems.
We evaluate the proposed method on the multilingual LAMBADA dataset, WMT24 dataset and the SST-2 benchmark,
demonstrating both its resilience to orthographic noise and its effectiveness in multilingual settings.

% The autoregressive language models suffer from orthographic attacks when the input text is noised by characters from multilingual alphabets, leading to a significant drop in performance. 
% This is caused by the out-of-vocabulary problem related to the tokenizer and the embedding look-up table.
% We introduce the pixel approach into the autoregressive language model to transfer the text-based embedding space into the pixel-based embedding space, which is more robust against the noise, by rendering the words into images.
% We conducted the experiments on the LAMBADA multilingual dataset and GLUE SST-2 benchmark to illustrate the robustness against the noise and the compatibility of the multilingual environment.
% autoregressive language model. 

\end{abstract}

\section{Introduction}
% Salesky: OOV + Tokenizer, Human -> visual embedder 
% Rust: pml + oov + tokenizer (finite vocabulary) 
% PDF: historic docs + why hard + several old solutions -> pixel
% B: dialect + tokenizer + LM 
% PIXAR: tokenizer + con (fix vocab) -> OOV  
% 
% here 
% OOV -> one solution: tokenizer -> better solution: pixel
% -> pixel + tokenizer free hard compatible with generative LM (next token prediction)
% -> ours
In real-world scenarios, textual data often deviates from standardized forms. 
Typographical errors occur unconsciously, while in other cases, people deliberately modify spellings as part of cultural or stylistic practices. 
In multilingual environments, speakers frequently mix words from different languages, or employ dialectal variations with unconventional orthography that remain intelligible to others. 
Additional sources of noise include errors introduced by tools such as Optical Character Recognition (OCR). 

Such diversity poses significant challenges for language models.
Models trained on monolingual and standardized corpora tend to suffer from performance degradation due to the out-of-vocabulary problem, as the space of possible variations is effectively unbounded \cite{oov2}. 
A practical approach optimizes the tokenization strategies, breaking words into smaller subwords to improve coverage \cite{tok1, tok2, tok3, BPE}. 
While subword tokenization mitigates some issues, it struggles with orthographic noise because the finite vocabulary cannot capture infinite variations \cite{tokenizer-attack}.

Human readers, however, demonstrate notable tolerance in interpreting noisy, multilingual, or dialectal text \cite{humaneye, pixel-salesky}. 
Inspired by this, pixel-based language modeling has emerged as a promising approach. Here, the text is first rendered as an image, and the model processes the resulting visual representation. 
This tokenizer-free paradigm eliminates dependence on a fixed vocabulary, leading to greater robustness against noise and variation. 
Pixel-based encoders have shown substantial benefits for understanding tasks.

However, this tokenizer-free structure introduces incompatibilities with generative language modeling. 
Since input units are image patches rather than aligned tokens, models lack explicit targets for next-token prediction, making it difficult to compute training losses. 
Some recent studies have attempted to bridge this gap. 
For instance, \citeposs{PIXAR} method based on Visual Transformer \cite{ViT} reformulates text generation as next-patch prediction, where the model generates image patches corresponding to the next word. 
Yet, these approaches require an additional OCR step to convert generated patches back into text, which introduces complexity and potential errors. 
Moreover, producing coherent and semantically meaningful next words through image generation remains a fundamental challenge.
Other recent work has explored a multimodal approach \cite{PixelGPT}, where the model is jointly trained on paired text tokens and corresponding image patches. 
In this setting, the model performs next-patch prediction for images and next-token prediction for text simultaneously, leading to stronger performance on downstream tasks. 

Nevertheless, a direct and efficient bridge from pixel-based input to text-based output is still lacking. 
This gap limits the seamless integration of pixel-based methods into the broader NLP ecosystem. 
In this work, we propose a pixel-based approach in generative language modeling in which each word is rendered as a separate image with a fixed resolution and a scalable font size, rather than rendering the entire sentence as a single image. 
This establishes a direct one-to-one correspondence between words and their pixel representations, 
thereby aligning the pixel-based input with the next-token prediction objective. 
As a result, our method naturally supports text generation without requiring any additional components such as OCR.

Furthermore, we observe that similar to other pixel-based methods, the pixel embedding space derived from images also demonstrates greater robustness to orthographic noise and stronger adaptability to multilingual text than the text-based methods which are constrained by symbolic token vocabularies. 

% This advantage arises from the fact that the representations are learned from visual features rather than being constrained by symbolic token vocabularies.

The remainder of this paper is organized as follows: in Section \ref{method}, we present our method, including the renderer, its integration with generative language models, and the acceleration strategy. 
And section \ref{experiment} reports experimental results that highlight the compatibility and robustness of our approach in handling noisy and multilingual text. 

\section{Related Works}

Pixel-based approaches render text into images and process the resulting visual representations instead of relying on symbolic tokenizer to process the input text. 
Existing studies vary in both how textual information is extracted from rendered images and how these representations are integrated into language model architectures.

Early work by \citet{pixel-salesky} et al. renders an entire sentence as an image and processes it through sliding windows and convolutional neural network \cite{CNN} blocks to obtain embeddings. 
These embeddings are then fed into a standard encoder–decoder transformer for machine translation tasks.
This work shows potential for the robustness of the pixel-based methods against various types of noise. 

\citeposs{pixel-rust} work PIXEL builds in the style of Masked Autoencoders (MAE) \cite{Masked-Autoencoder}. 
In these models, texts are rendered into a single image, which is split into fixed-size patches. 
Each patch is projected into a vector representation through a linear layer. 
An autoencoder is trained to reconstruct masked patches, and the encoder is subsequently fine-tuned for downstream tasks. This method shows a good understanding of language and strong robustness to noise. 

Extensions of PIXEL include applications to historical document understanding \cite{PHD}, where image sizes are increased and training incorporates both artificial and real scanned data, 
and the application of 
dialectal language modeling \cite{pixel-barbara}, where visual representations capture better various spellings in dialects that are challenging for text-only models.

Beyond encoder-based frameworks, some studies have explored adapting this style to decoder architectures. 
For instance, PIXAR \cite{PIXAR} claims to be the first transformer-decoder pixel-based model. 
It renders texts as a single image, splits it into patches, and autoregressively reconstructs these patches. 
Training involves two stages: autoregressive reconstruction followed by Generative Adversarial Networks \cite{GAN} fine-tuning to improve the readability of generated image patches. 
Although it shows the effectiveness in robustness against noise, this approach still requires an auxiliary OCR module to convert generated patches into text, which limits its practicality for generative tasks.

PixelGPT \cite{PixelGPT} is another work in this line.
It combines image patches with text tokens in a multimodal decoder. 
The model is trained with dual prediction heads: one for next-patch prediction for the image input and another for next-token prediction for the text modality input. 
Despite this multimodal design, its evaluation largely focuses on downstream classification tasks rather than deeply addressing text generation.

Although these approaches demonstrate the promise of pixel-based representations, they share a common limitation: text is rendered into a full image and then partitioned into fixed-size patches.
This design leads to two fundamental problems. 
First, individual characters or punctuation marks may be split across patches, resulting in fragmented and incoherent outputs during generation. 
Second, patch boundaries do not align with linguistic units such as words, making next-token prediction infeasible and complicating the integration of pixel-based models into generative language modeling.

To overcome these limitations, we propose a simple yet effective strategy: rendering each word as an individual fixed-size image rather than slicing entire sentences into patches. 
This ensures that every word is preserved as a whole unit, enabling direct alignment between pixel-based inputs and next-token prediction. 
Our method retains the robustness and generalization ability of pixel-based models, while seamlessly supporting text generation without requiring additional modules such as OCR.

In parallel, \citeposs{pixel2025} recent work has attempted to combine pixel-based representations with pre-trained Large Language Models (LLMs) for tasks such as machine translation. 
Particularly, prompts remain in plain text while source-language tokens are split into bigrams and rendered as image slices, which are then aggregated through average pooling into word-level pixel embeddings. In contrast, our approach focuses on addressing the weaknesses of prior pixel-based generative models. By adopting a transformer decoder structure and training with next-token prediction, we demonstrate that our pixel-based method achieves robustness to noise while maintaining strong adaptability to multilingual text.

\section{Method}
\label{method}
% Our method focuses on the solution combining the pixel method with generative transformer decoder. We introduce here the Renderer adaptive for the generative task, as well as our model architecture, as well as the strategy to accelerate the renderer. 

Our method centers on integrating the pixel-based approach with a generative Transformer decoder. In this section, we introduce the adaptive renderer designed for generative tasks, present our strategy for accelerating the rendering process, and describe the overall model architecture.

\begin{figure}
    \centering
\includegraphics[width=1.1\linewidth]{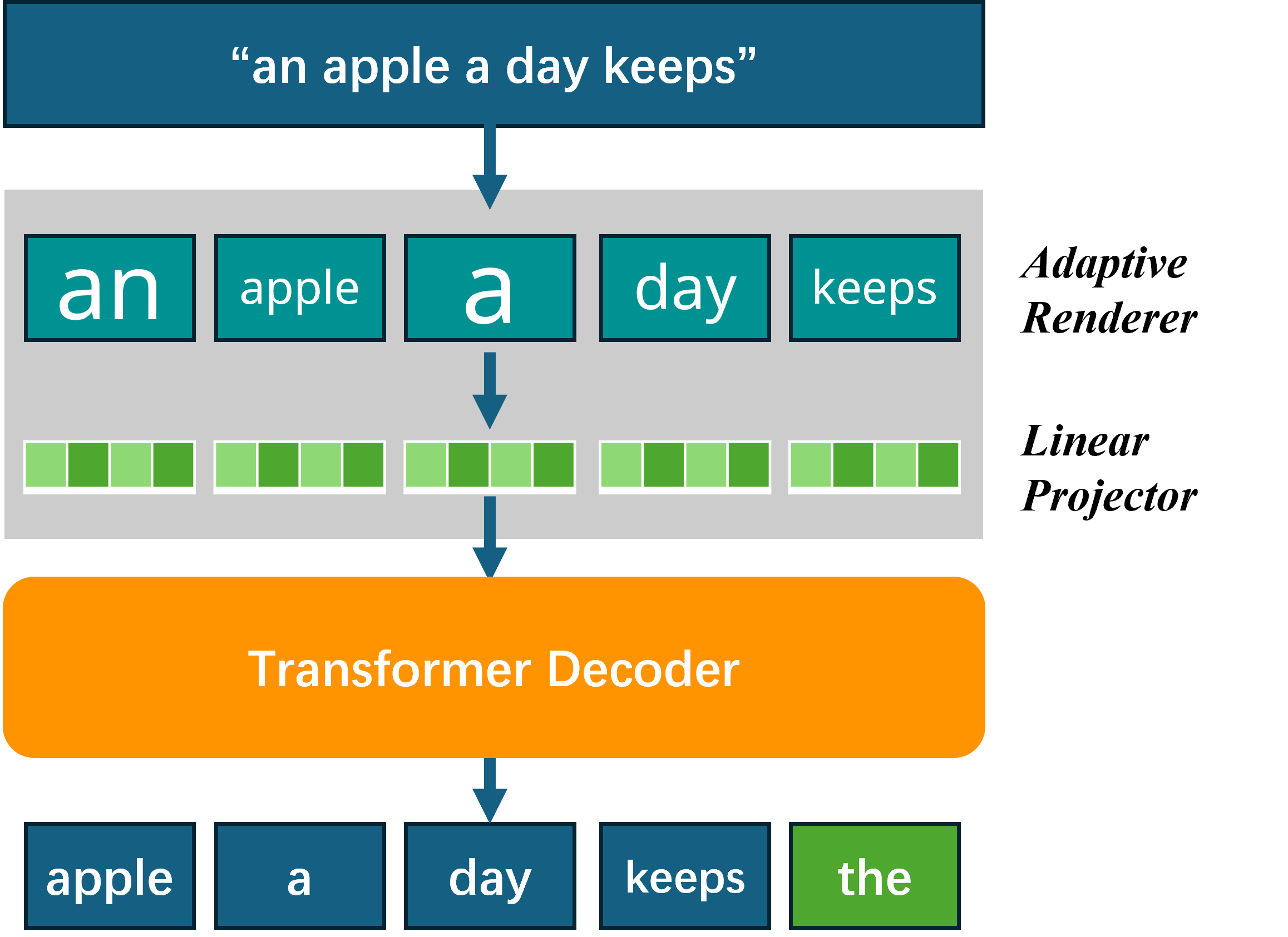}
    \caption{Our pixel-based generative language model is built on top of a Transformer decoder and additionally incorporates an adaptive renderer and a linear projector. The adaptive renderer converts input words into images, which are subsequently mapped into pixel embeddings. 
    % The decoder processes these pixel representations and autoregressively predicts the next tokens
    }
    \label{fig:model}
\end{figure}

\subsection{Adaptive Renderer}

Unlike previous pixel-based methods that render the entire input text as a single image and making it incompatible with the natural training objective of generative language models. 
We propose an optimized rendering strategy where each word is rendered into a separate image.

To account for word length, we adaptively adjust the font size: shorter words are scaled up, while longer words are scaled down. 
This ensures that all words can be uniformly represented within images of fixed width and height, 
making them directly transferable into embedding representations.

% When integrated with a tokenizer, the sentence is first tokenized into a sequence of tokens with corresponding IDs. 
% Each token is then rendered into an image, which serves as the model input, while the token ID provides the prediction target.

% Different from the previous contributions in the pixel methods, where the complete input text is rendered in one image, and it is not able to be compatible with the natural training task, next token prediction, of a generative language model, we optimized the strategy of the rendering, where each word is rendered into a separate image. Further, considering the word length, we scale the font size with a ratio: 

% A shorter word will be scaled up, while a longer word will be scale down. 

% So that all words can be rendered adaptively in an image with fixed width and length, and are possible to be transferred to an embedding representation. 

% When compatible with a tokenizer, the sentences are tokenized with a sequence of tokens and corresponding ids, we can take the sequence of the images of the tokens as input, while take the token id as the target. 

\subsection{Acceleration Strategy}

A key challenge compared with prior pixel-based approaches lies in rendering efficiency. 
Previous methods render the entire input text once as a single image, whereas our method requires rendering each word individually, resulting in a complexity of $\mathcal{O}(n)$ during both training and inference.

To address this, we pre-render all tokens from the tokenizer’s vocabulary and store them in a dictionary. 
Rendering a token then reduces to a simple dictionary lookup with complexity $\mathcal{O}(1)$. 
This lookup mechanism can be further optimized by implementing it as an embedding layer in PyTorch.
Consequently, rendering a sequence of tokens is equivalent to multiplying a vector of token indices $(s, |v|)$ 
with a precomputed matrix $(|v|, \text{width} \times \text{height})$.

% For stability, the renderer embedding layer is kept frozen during training.

% A challenge compared with the previous pixel-related works is that the previous work renders the input text only once in one image, but we should render each word with complexity $\mathcal{O} (n)$ during the training or inference. 

% As an acceleration strategy, we render the tokens from the vocabulary of the tokenizer in advance and form them as a dictionary, and we replace the rendering by looking up from the dictionary with a complexity of $\mathcal{O} (1)$. 
% The look-up dictionary can be further accelerated technically  as an embedding layer in the PyTorch, where the embedding layer is essentially a look-up table. Rendering a sequence of tokens is then simplified as the multiplication between a vector with a matrix: $(s,|v|) x (|v|, width \times height)$.
% We set the renderer embedding layer frozen during the training. 

\subsection{Pixel Generative Language Model}

We build our model upon the LLaMA architecture \cite{LLaMA-1} by replacing its original embedding layer with a pixel embedding layer, as shown in Figure \ref{fig:model}.

This layer consists of two components: (1) adaptive renderer, a pre-rendered lookup embedding that efficiently maps tokens to their corresponding word images, and (2) linear projector: a learnable linear projection that maps each image into a fixed-size representation required by the transformer decoder blocks.

Concretely, an input sequence is first converted into a sequence of images via the renderer embedding layer. 
Each image is then projected through the linear layer to obtain dense, continuous pixel embeddings of uniform size. 
These embeddings serve as the inputs to the transformer decoder, which proceeds unchanged from the original LLaMA model. 
Hidden states are computed at each position, and the output head predicts the next token ID in the sequence. 
Thus, our model establishes a mapping of tokens through the image space and predicts the next tokens.
% , i.e., 
% $t \;\rightarrow\; I \;\rightarrow\; e,$
% where $t$ denotes tokens, $I$ their rendered images, and $e$ the pixel embeddings.

% We refer to this architecture as a Pixel Generative Language Model. 
This architecture receives inputs in the pixel modality while remaining fully compatible with next-token prediction. 

\subsection{Training and Inference}

During training, a sentence is first tokenized into a sequence of token IDs 
$\mathbf{t} = (t_1, \dots, t_n)$. 
Each token $t_i$ is mapped to its corresponding image $I(t_i)$ via the renderer, 
and then projected by the linear layer into a pixel embedding $\mathbf{e}_i$. 
The resulting sequence $(\mathbf{e}_1, \dots, \mathbf{e}_n)$ serves as input to the transformer decoder. 
The prediction head outputs the probability distribution over the next token at each position:
\[
P(t_{i+1} \mid I(t_i), \dots, I(t_0)), 
\]
and the model is trained with the standard cross-entropy loss over the token sequence.  

During inference, if the input contains tokens not present in the pre-rendered vocabulary, 
these tokens can still be rendered individually into images. 
Alternatively, an extended lookup table can be constructed like a patch by pre-rendering additional words.
In both cases, the OOV tokens can be obtained as in-vocabulary tokens in the pixel embedding space, 
i.e. when the input text is noisy, the token sequence distribution changes, i.e.,
\[
P(t_1, \dots, t_n) \;\neq\; P(t_1', \dots, t_m'),
\]
where $(t_1', \dots, t_m')$ denotes the corrupted sequence. 
However, the rendered images of the two sequences remain visually similar, leading to
\[
P(I(t_1), \dots, I(t_n)) \;\approx\; P(I(t_1'), \dots, I(t_m')),
\]
which explains the robustness of our pixel-based approach to noise and orthographic variation. 

% For sequence generation, each predicted token is first rendered into an image using the renderer, 
% then projected into its pixel embedding, which is fed back into the model for subsequent prediction.  

% During the training, we receive the token ids tokenized by the tokenizer from the sentence, while we transfer them to the corresponding images, through the linear layer, we obtain their pixel embeddings. 
% The rest of the transformer and prediction header predicts the next token at each position. $P(t_i+1, I_i)$. Similar as the next-token prediction, we calculate the cross entropy loss, with the sequence of tokens. 

% During the inference, even if the sentence contain the tokens which are not in the tokenizer, we can still render the out-of-vocabulary tokens separately, or through an additional look-up table if we collect those words and render them before using. We can still obtain the pixel embeddings of the OOV tokens.

% During the generation of a sequence of tokens, a predicted token is mapped to the pixel embedding through the intermediate image rendered by any variations of the renderer.

% When a text is noised, the tokens are then different, and the do not follow the original distribution, namely $P(t1..tn) \not = P(t1’ … tn’)$. However visually the words or subwords are visually similar $P(I(t1)..I(tn)) wave = P(I(t1’) … I(tn’))$.

\section{Experiment}
\label{experiment}
% Previous pixel-based works render an entire sentence as a single image, 
% and therefore do not require a tokenizer as part of the input pipeline. 
% In contrast, our approach targets \emph{generative language modeling} with next-token prediction. 
% This setting requires a tokenizer in order to assign token IDs as training targets and to compute the cross-entropy loss.  

In our experiments, we aim to answer the following research questions:

\begin{itemize}
    \item[\textbf{Q1}] How well does our approach perform in multilingual settings, 
    where the input may contain multiple languages?
    \item[\textbf{Q2}] How robust is our pixel-based generative language model to noise, 
    compared with a standard text-based generative language model?
\end{itemize}

To answer these questions, we pretrain our pixel language model and evaluate it on the multilinguality and robustness in benchmarks.
% the previous works are all render a whole sentence as an image, thus a tokenizer is for the input not neccssary

% however we focus on the generative language model with next-token prediction,
% thus we need a tokenizer to assign the token ids as the target during the training to calculate the loss.

% We focus on x research questions. Q1: 
% Q1: how is the robustness against the noise compare to a text-based generative language model
% Q2: how is the multilingual performance

\subsection{Pretrain}

% Pretrain

We pretrain our model on the same datasets as PIXAR~\cite{PIXAR}, namely 
BookCorpus~\cite{Bookcorpus} and English Wikipedia. 
To ensure comparability, we adopt the identical preprocessing pipeline: 
sentences shorter than 100 characters or longer than 800 characters are removed, 
and the preprocessing tool provided by PIXAR is applied.  

For tokenization, we use the Byte-Pair Encoding (BPE, \citealp{BPE}) tokenizer from the LLaMA model, 
trained on the pretraining corpus with a vocabulary size of 32,001. 
Furthermore, prior to training, we render the entire vocabulary into word-level images, 
which are then used as the basis for constructing our pixel embedding space.  

We compare our approach against a baseline LLaMA model \cite{LLaMA-1} equipped with a standard 
token-based embedding layer. 
Both models are designed to have comparable parameter sizes and are trained 
on the same pretraining datasets for fairness. Both model are trained by 100K steps on a single NVIDIA A40 for 5 hours. We show the pretraining details in Table \ref{tab:pretrain}. 
% For evaluation, we report the same set of metrics on multilingual performance, 
% as well as the corresponding results on robustness to noise and downstream tasks.  

% We apply the same datasets used for the pretraining in work PIXAR \cite{PIXAR}, namely BookCorpus \cite{Bookcorpus} and Wikipedia\footnote{\url{https://en.wikipedia.org/}}. We also implemented the same preprocessing method. We filtered out all sentences shorter than 100 characters and larger than 800 characters, and used the same preprocessing tool provided by the baseline work PIXAR. 

% We used the same BPE tokenizer as in the Llama model, we train it on the pretraining dataset with 32001 tokens. We also render vocabularies list each as an image before our pretraining. 

% Multiliguality 

\begin{table}[]
\centering
\resizebox{0.95\columnwidth}{!}{%
\begin{tabular}{l|cc}
\hline
 & \multicolumn{1}{c|}{PM} & LM \\ \hline
font & \multicolumn{1}{c|}{Go Noto Current} & - \\ \hline
basic font size & \multicolumn{1}{c|}{10} & - \\ \hline
image height & \multicolumn{1}{c|}{20} & - \\ \hline
image weight & \multicolumn{1}{c|}{50} & - \\ \hline
image channel & \multicolumn{1}{c|}{1} &  \\ \hline
hidden size & \multicolumn{2}{c}{768} \\ \hline
intermediate size & \multicolumn{2}{c}{2048} \\ \hline
num attention heads & \multicolumn{2}{c}{12} \\ \hline
num hidden layers & \multicolumn{2}{c}{12} \\ \hline
vocab size & \multicolumn{2}{c}{32001} \\ \hline
max position embeddings & \multicolumn{2}{c}{2048} \\ \hline
rms norm eps & \multicolumn{2}{c}{1.00E-05} \\ \hline
\end{tabular}
}
\caption{Hyperparameters of pixel model (PM) and language model (LM)}
\label{tab:pretrain}
\end{table}

\begin{table*}[h]
\centering
\resizebox{2.2\columnwidth}{!}{%
\begin{tabular}{l|r|r|r|r|r|r|r|r}
\hline
   & \multicolumn{1}{c|}{en} & \multicolumn{1}{c|}{de} & \multicolumn{1}{c|}{fr} & \multicolumn{1}{c|}{it} & \multicolumn{1}{c|}{ru} & \multicolumn{1}{c|}{zh} & \multicolumn{1}{c|}{ja} & \multicolumn{1}{c}{hi} \\ \hline
LM & 495                     & 28857 (58)              & 14064 (28)              & 19102 (39)              & 1302480 (2627)          & 424341 (856)            & 503817 (1016)           & 228299 (461)            \\ \hline
PM & 279                     & 22387 (80)              & 8478 (30)               & 12918 (46)              & 298420 (1068)           & 7606 (27)               & 5578 (20)               & 6920 (25)               \\ \hline
\end{tabular}
}
\caption{Result of perplexity of our pixel model and language model (LM) on WMT24 dataset in different languages including English (EN), German (de), French (fr), Italian (it), Russian (ru), Chinese (zh), Japanese (ja) and Hindi (hi). We report the absolute values for all languages, and for non-English languages provide relative increase w.r.t English for each model}
\label{tab:result-wmt}
\end{table*}

\begin{table*}[h]
\centering
% \resizebox{1.2\columnwidth}{!}{%
\begin{tabular}{c|r|r|r|r|r}
\hline
 & \multicolumn{1}{c|}{en} & \multicolumn{1}{c|}{de} & \multicolumn{1}{c|}{es} & \multicolumn{1}{c|}{fr} & \multicolumn{1}{c}{it} \\ \hline
LM & 269 & 20195 (75) & 14367 (53) & 14360 (53) & 16390 (61) \\ \hline
PM & 138 & 16858 (121) & 7909 (57) & 8554 (62) & 10215 (74) \\ \hline
\end{tabular}
% }
\caption{Result of our pixel model (PM) and the language model (LM) on multilingual LAMBADA dataset in different languages including English (en), German (de), Spanish (es), French (fr) and Italian (it). We report both absolute value and relative increase w.r.t English for each model}
\label{tab:result-LAMBADA}
\end{table*}

\subsection{Multilinguality}

We evaluate token-level perplexity on two multilingual benchmarks: 
the multilingual LAMBADA dataset~\cite{LAMBADA, radford2019language} and WMT24 \cite{WMT24}. 
For LAMBADA, we use the official test set, which consists of English (en) sentences 
along with their translations in German (de), Spanish (es), Italian (it), and French (fr).  
From the WMT24 dataset, we additionally sample 100 English sentences with lengths between 200 and 300 characters, along with their parallel translations in German, Italian, French, and in non-Latin languages including Russian (ru), Chinese (zh), Japanese (ja), and Hindi (hi).  We report the token-level perplexity for each language, where higher values indicate greater difficulty or confusion in modeling sentences of that language.

Table \ref{tab:result-LAMBADA} summarizes the results on the multilingual LAMBADA dataset. 
Although the pixel model achieves lower absolute perplexity than the language model in English, it exhibits a much larger relative increase when extended to other Latin-alphabet languages. 
In German, Spanish, French, and Italian, the perplexity of pixel model grows substantially more than that of the language model, 
indicating that the language maintains better cross-lingual generalization in languages closely related to English.

Table \ref{tab:result-wmt} shows result in further languages.
For Latin-alphabet languages, the pixel model follows the same pattern observed in the LAMBADA experiment, showing larger increases in perplexity than the language model. 
In contrast, the trend reverses for non-Latin languages: the pixel model proves substantially more robust, with relative increases several orders of magnitude smaller than those of the language model in Russian, Chinese, Japanese, and Hindi. 
These findings indicate that while the language model generalizes more effectively across languages related to English, the pixel model exhibits superior resilience in languages written in non-Latin scripts.

The result reveal a clear divergence in the relative performance of the pixel model and the language model depending on the linguistic family and writing system. 
In Latin-based languages, pixel model underperforms compared to LM. 
This discrepancy can be attributed to the partial compatibility of these languages with the English tokenizer employed in our experiments. 
Many lexical items in Latin-alphabet languages share cognates or morphemic units with English, 
enabling language model to effectively reuse subword units across languages.
% Subword segmentation further alleviates the challenges posed by rich inflectional morphology, 
% such as conjugation and declension patterns, thereby reducing the learning burden on LM. 
% In contrast, PM, which operates on lower-level representations such as characters or pixels, 
% must reconstruct higher-level linguistic structures from raw input. 
% In scenarios where tokenization is already effective, this design choice limits its efficiency and results in higher perplexity compared to LM.

By contrast, in non-Latin writing systems languages, the pixel model demonstrates a substantial advantage over language model. In these languages, the English tokenizer fails to provide meaningful subword units: in Russian, the Cyrillic alphabet is not representative in the English dataset, while in logographic or syllabic systems such as Chinese and Japanese, individual characters are often treated as independent tokens without semantic composition. The pixel model captures better visual clues than the language model. 
% As a result, the language model exhibits abnormally high perplexity in these settings, reflecting its reliance on subword abstractions that do not transfer well across writing systems. 
% The pixel model, however, is less dependent on tokenization and instead captures visual aspects.

% , enabling it to achieve dramatically lower perplexities. For example, while LM’s perplexity in Chinese and Japanese exceeds 800 and 1000 respectively, PM reduces these values to below 30, highlighting its robustness in languages where tokenization is inadequate.

% Taken together, these findings demonstrate that the relative performance of PM and LM is strongly conditioned by the interaction between language characteristics and tokenizer alignment. LM excels in Latin-based languages where subword overlap with English mitigates morphological complexity, while PM exhibits superior adaptability in non-Latin or highly inflected languages where tokenization is ineffective.

\begin{figure*}[h]
  \includegraphics[width=0.45\linewidth]{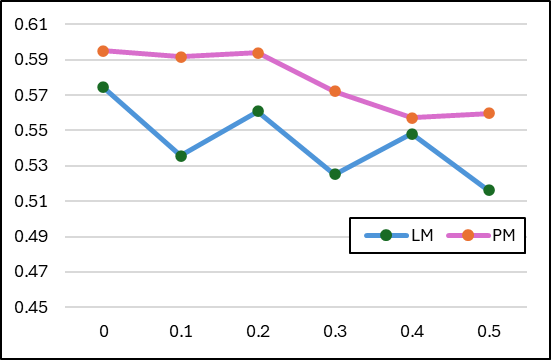} \hfill
  \includegraphics[width=0.45\linewidth]{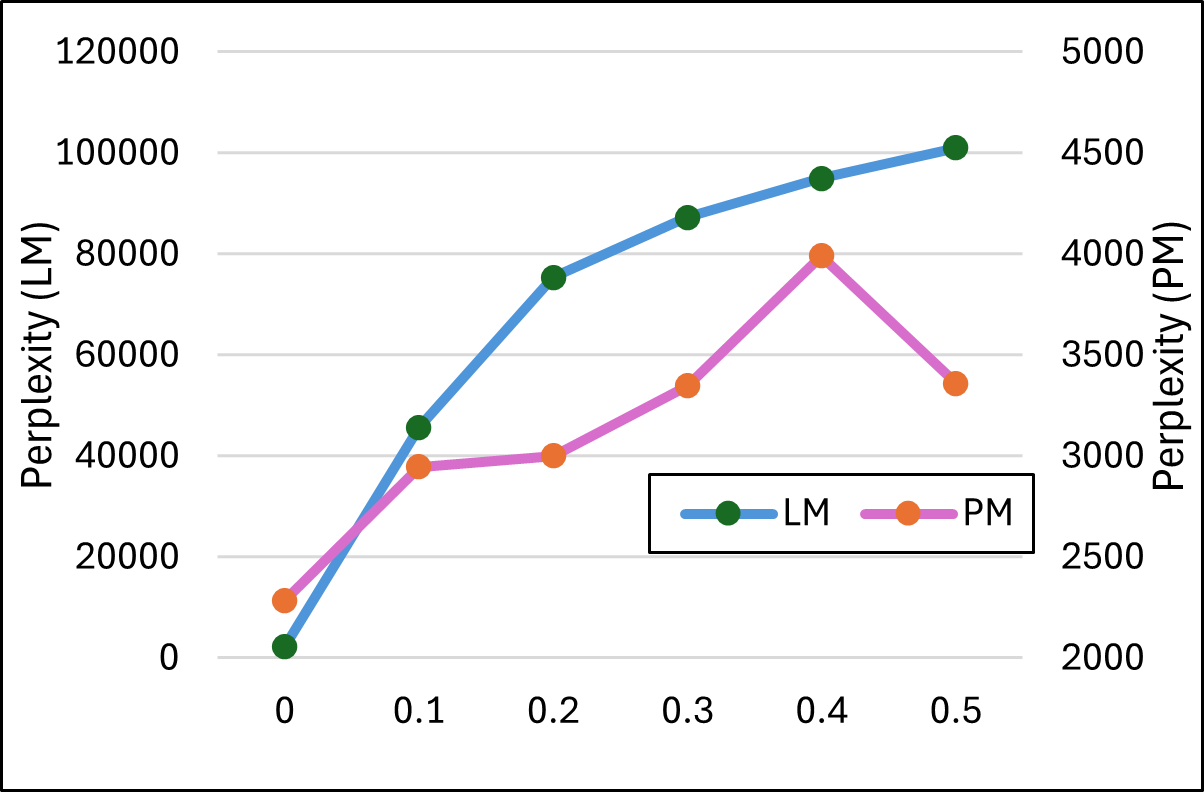}
  \caption {Accuracy (left) and perplexity (right) of the pixel model (purple) and the language model (blue) on the SST-2 dataset}
  \label{fig:result-sst2-acc-ppl}
\end{figure*}

\subsection{Robustness against Noise}

To evaluate robustness under noisy conditions, we follow the setup in PIXAR. 
Specifically, we inject noise into sentences from the English LAMBADA dataset by replacing 
each character with a certain probability. 
The replacement character is randomly sampled from a predefined noise dictionary
as introduced in PIXAR. 
We then report the change of token-level perplexity as the noise ratio increases. We further assess robustness on a downstream classification task, namely 
Stanford Sentiment Treebank (SST-2) \cite{SST2}. 
We finetune our model on the train set by attaching a prediction head that receives the last hidden state 
from the final decoder layer, while keeping the rest of the model frozen. 
We report accuracy, precision, recall, and token-level perplexity.

% Finally, we provide an ablation study in Appendix~\ref{...}, 
% where the prediction head instead takes as input the average of the hidden states 
% from the final decoder layer.  

% To illustrate the degrade of the performance with increasing ratio of noise, we follow the experiments on the PIXAR \cite{PIXAR}, where we insert certain ratio of noise in the sentence from the english LAMBADA dataset. Namely each character has a probability to be replaced by the noised character. The noised character is random selected from a pre-defined dictionary, as defined in the PIXAR. We report the change of the token-level perplexity.

% We further evaluate the robustness against the noise on the downstream task, namely the Stanford Sentiment Treebank (SST-2). We finetune our model with a prediction head, where the head receives the last hidden state on the last decoder layer. Our model remains frozen during the finetuning. We report the accuercy, along with the precision, recall, as well as the token-level perplexity. 

% We also show the ablation study in Appendix \ref{}, where during the finetuning we average the last hidden states as the input feature of the prediction head.

\begin{table}[]
\resizebox{1\columnwidth}{!}{%
\begin{tabular}{l|r|r|r|r|r|r}
\hline
   & 0   & 0.1   & 0.2   & 0.3   & 0.4   & 0.5   \\ \hline
LM & 269 & 18581 & 46543 & 63840 & 73288 & 79457 \\ \hline
PM & 139 & 185   & 242   & 311   & 393   & 485   \\ \hline
\end{tabular}
}
\caption{Perplexity measured on the LAMBADA with increasing noise}
\label{tab:result-LAMBADA-noise}
\end{table}

\begin{figure}
    \centering
    \includegraphics[width=0.9\linewidth]{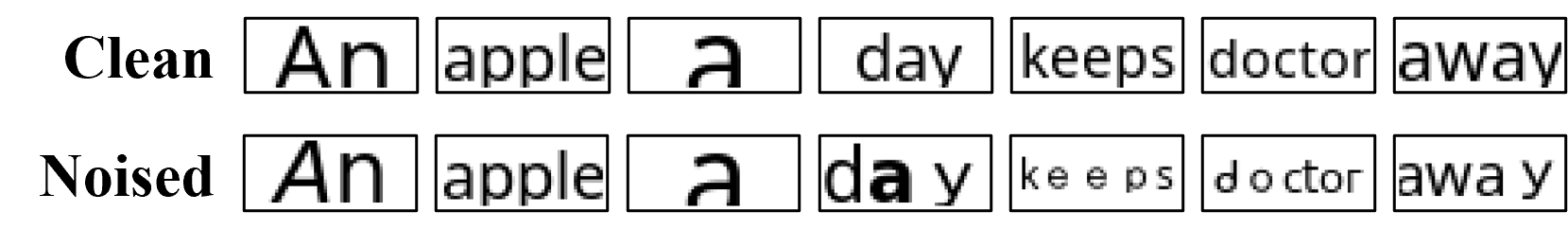}
    \caption{Showcase of rendered images of a sentence ``\textit{an apple a day keeps doctor away}". The noised sentence cosine-similarity between the sentence and the noised sentence is 0.89 in the pixel embedding space}
    \label{fig:showcase}
\end{figure}

% Figure \ref{fig:result-sst2-acc-ppl} (left) presents the accuracy of the sentiment classification task. Overall, both models exhibit decreasing accuracy as the noise ratio increases.
% accuracy decreases with increasing ratio of noise, while pixel model drops by 0.3, while language model drops by 0.6 in total. 
% The precision for pixel model decreases by 0.4, and the recall changes by 0.2, while the language model decreases by 0.5 in precision and 0.6 in recall. 

Figure \ref{fig:result-sst2-acc-ppl} (left) presents the accuracy results for the sentiment classification task. As the noise ratio increases, both models show declining performance. Overall, the pixel model experiences a smaller accuracy drop (0.3) compared to the language model (0.6). A similar pattern is observed for precision and recall: the pixel model decreases by 0.4 and 0.2, respectively, whereas the language model shows larger drops of 0.5 in precision and 0.6 in recall. These results indicate that the pixel model maintains more stable performance under noisy conditions.
Additionally, Figure \ref{fig:result-sst2-acc-ppl} (right) reports perplexity under noisy conditions. The perplexity of the language model increases by a factor of 43 compared to the clean setting, reaching an absolute value of 98,692, whereas the pixel model increases only by 1.4 times, corresponding to 1,073 in absolute terms. A similar trend is observed on the noisy LAMBADA dataset (Table \ref{tab:result-LAMBADA-noise}), where the perplexity of the pixel model grows by at most 3.49 at the highest noise level, while the language model exhibits a dramatic increase of 294 times.
% The perplexity of the language model increases by 43 times compared to the clean setting and reaching an absolute value of 98,692, whereas the pixel model increases only by 1.4 times, corresponding to 1,073 in absolute terms.
% A similar trend is observed on the noisy LAMBADA dataset, as shown in Table \ref{tab:result-LAMBADA-noise}, where the perplexity of the pixel model increases by at most 3.49 under the highest noise level, while the language model shows a dramatic increase of 294 times.

These results indicate that the pixel model is substantially more resilient to noise compared to LM. 
The difference can be attributed to model design: LM depends on subword tokenization, which is highly sensitive to corrupted character sequences,
consequently the noisy sentence is interpreted as a different sequence of tokens.
The pixel model weakens the dependency on tokenization,
especially even a word is perturbed, it can still be rendered as a clear image, and maintains the visual similarity in the pixel embedding space. 
We demonstrate an example by showcasing an example ``\textit{an apple a day keeps doctor away}'' and its noised sentence as shown in Figure \ref{fig:showcase}. Their sentence level cosine-similarity is 0.86 in the pixel embedding space, while the letter ``A'' in word \textit{an}, ``a'' in \textit{day}, ``e'' in \textit{keeps}, ``d'', ``o'', ``c'' in \textit{doctor}, ``a'' and ``y'' in \textit{away} are replaced, but the rendered images still maintains the visual similarity.

\section{Limitation}

% assumption

In our robustness experiments against orthographic noise, the performance of the pixel 
generative language model is evaluated under the assumption that the tokenization of a noised 
sentence remains identical to that of the original clean sentence.  

In practice, however, introducing noise inevitably alters the tokenization in most cases. 
Once a sentence is re-tokenized, noisy tokens might be split into different subwords.
Since our primary focus is to illustrate the benefit of the pixel method in generative language modeling, the pixel embedding space still retains the visual similarity, thus we adopt this simplifying assumption. 
We regard the re-tokenization of noised sentences for the token-level pixel methods as an open problem, 
and leave its resolution for future work.  

% On the experiments for the robustness against the orthographic noise, the performance of our pixel generative language model is based on an assumption, that a tokenization of a noised sentence remains the same as the sentence without noise. 

% Theoretically the noise in sentences leads to a different tokenization unavoidablly in the most cases. 
% But once the noised sentence is re-tokenized, a noised token is then splitted.
% In this work we mainly focus to introduce the benefit of the pixel embedding space brought by the visual similarity, 
% while the re-tokenization leads to a loss of the visual similarity. 
% % and the tokenization is in fact not a part of it. 
% % with the lost of the visual similarity. 
% To show the potential benefit of robustness against noise of pixel method on the generative language model, we conduct our experiment on an assumption, 
% that the tokenizer of our pixel language model performs stable on the nosied sentences as when they are not being noised
% % a tokenization of a noised sentence remains the same as the sentence without noise.
% concretely the number of the tokens stays the same, our pixel model takes only the subwords as the input but not the token ids. 
% % while the token ids of the subwords   
% We choose to keep the re-tokenization of the noised sentence  as an open task, and will solve it in the future. 

\section{Conclusion}

In this work, we proposed a pixel-based generative language model that bridges 
the gap between pixel representations and next-token prediction. 
Unlike prior approaches that render entire sentences into images and rely on patch-based 
representations, our method renders each word as an individual image of fixed size, 
which naturally aligns with token-level prediction. 
We further introduced an adaptive renderer that scales font size according to word length, 
and an efficient acceleration strategy that replaces rendering with a pre-rendered 
lookup table implemented as an embedding layer. Through experiments on multilingual benchmarks and noisy text scenarios, 
we demonstrated that our model achieves stronger robustness against orthographic noise 
and improved adaptability in multilingual settings, especially non-Latin languages. 

\section*{Acknowledgments}
Han Yang received funding from the Deutsche Forschungsgemeinschaft (DFG) under grant number: MA 3964/15-3 (SocioHub project).

\bibliography{custom}

% \appendix

% \newpage

% \appendix

% \section{Detailed Result}

% We provide here more detailed results from the experiments.
% In Table \ref{tab:app-mlambada} we show the absolute value of the perplexity of the pixel model (PM), and the baseline language model (LM) in the multilingual LAMBADA dataset. We report the result of the absolute value of perplexity in Table \ref{app:result-wmt}.

% \begin{table}[]
% \begin{tabular}{l|r|r|r|r|r}
% \hline
%  & \multicolumn{1}{c|}{en} & \multicolumn{1}{c|}{de} & \multicolumn{1}{c|}{es} & \multicolumn{1}{c|}{fr} & \multicolumn{1}{c}{it} \\ \hline
% LM & 269 & 20,196 & 14,367 & 14,360 & 16,390 \\ \hline
% PM & 139 & 16,859 & 7,909 & 8,555 & 10,216 \\ \hline
% \end{tabular}
% \caption{Absolute value of perplexity measured on multilingual LAMBADA dataset}
% \label{tab:app-mlambada}
% \end{table}

% (Appendix: token length statistic, )
% One may concern that this we introduce the width and length as new hyperparameters, will some tokens not distinguishable if they are too long for the select length? 
% In fact during the training and testing, the model takes the input of tokens, tokenized by the tokenizer. We show the statistics here of the length of the vocabulary of an our BPE tokenizer, trained on the BookCorpus and English Wikipedia. We show here some example of the rendered images of words long enough. 

% We also show here rendering result of multilingual tokens. 

% \section{Example Appendix}
% \label{sec:appendix}

% This is an appendix.

\end{document}